# Online Semi-Supervised Concept Drift Detection with Density Estimation

**Chang How Tan**
Monash University,
Victoria, Australia 3800,
chang.tan2@monash.edu

**Vincent CS Lee**
Monash University,
Victoria, Australia 3800,
vincent.cs.lee@monash.edu

**Mahsa Salehi**
Monash University,
Victoria, Australia 3800,
mahsa.salehi@monash.edu

## Abstract

Concept drift is formally defined as the change in joint distribution of a set of input variables $X$ and a target variable $y$. The two types of drift that are extensively studied are *real drift* and *virtual drift* where the former is the change in posterior probabilities $p(y|X)$ while the latter is the change in distribution of $X$ without affecting the posterior probabilities. Many approaches on concept drift detection either assume full availability of data labels, $y$ or handle only the virtual drift. In a streaming environment, the assumption of full availability of data labels, $y$ is questioned. On the other hand, approaches that deal with virtual drift failed to address real drift. Rather than improving the state-of-the-art methods, this paper presents a semi-supervised framework to deal with the challenges above. The objective of the proposed framework is to learn from streaming environment with limited data labels, $y$ and detect real drift concurrently. This paper proposes a novel concept drift detection method utilizing the densities of posterior probabilities in partially labeled streaming environments. Experimental results on both synthetic and real-world datasets show that our proposed semi-supervised framework enables the detection of concept drift in such environment while achieving comparable prediction performance to the state-of-the-art methods.

## Introduction

Many real-world applications such as credit card fraud detection, mining of user interest, and network traffic monitoring rely heavily on data streams (Gao, et al., 2007). In these machine learning applications, typically the relations and patterns in data evolve over time which causes predictive learning models to become outmoded (Žliobaitė, et al., 2016). The challenge in learning from real-world domain is that the concept of interest depends on some hidden context which are usually uncaptured in the form of predictive attributes (Tsymbal, 2004). Often, the change of uncaptured hidden context in data attributes are the cause of concept drift which makes the learning task more complicated. An example of concept drift in traffic management is the usage behaviors of each road segment changes over time (Žliobaitė, 2010).

Concept drift refers to the change in joint distribution of the input variables $X$ and a target variable $y$ over time. In the context of machine learning, the target variable y is a label variable of a set of given features, $X$. Hence, studies of concept drift in machine learning context focus on how the given set of input variables $X$ affects the target variable $y$. In other words, concept drift researchers are generally interested in the change of distribution $X$, $p(X)$ and the change of distribution $y$ given $X$, $p(y|X)$. The change in $p(X)$ without affecting $p(y|X)$ refers to virtual drift while the change in $p(y|X)$ with or without change in $p(X)$ is referred as real drift (Gama, et al., 2014). Virtual drift does not capture the change in $p(y)$ (Delany et al. 2005), whereas real drift can only be detected with the availability of data labels.

Traditional online machine learning techniques deal with concept drift simply by learning incrementally from it. This is undesirable in some circumstances such as fraud detection, intrusion detection and online sentiment analysis (Dal Pozzolo, et al., 2015). Detecting concept drift is still required for a system to take appropriate reflection actions against drifts and hence become an essential component in stream learning.

Recent studies on concept drift detection are separated into supervised, semi-supervised and unsupervised areas. Supervised methods require data labels of all instances in order to compute performance-based measurement to sequentially monitor concept drift (Sethi & Kantardzic, 2017). However, the assumption of full availability of data labels is unrealistic as accessing label information might be expensive. Semi-supervised methods focus on learning and adapting well to data streams with the presence of concept drift and with only portion of data labels. Unsupervised methods focus more on detecting data distribution drift (Sethi & Kantardzic, 2017). Unsupervised methods assume no data labels are available in streaming environment.

In concept drift detection literature, predictive performance feedback is typically used to handle the real concept drift in supervised methods (Gama, et al., 2014). However, due to the limitation of data label availabilities in semi-supervised and unsupervised methods, they are unable to compute sequential performance measurement to detect real drift. The concern on these methods is that real concept drift is not efficiently resolved.

This paper proposes a generic semi-supervised framework to address the challenges above where real concept drift is addressed under a realistic streaming environment where there are little to no data labels. The proposed framework incorporates learning methods such as Active learning (Settles, 2009) and Positive Unlabeled (PU) learning (Bekker & Davis, 2018) to discover reliable labeled data (i.e. the estimated data labels inferred (or extracted) for the unlabeled data with high confidence) from the unlabeled portion in data streams. The posterior probabilities from the current reliable labeled data is compared to the posterior probabilities generated from an incremental estimator which learns incrementally from previous reliable labeled data obtained. Density estimation is used as a comparison method for these posterior probability distributions as statistical comparison methods are unstable for distributions which are partially labeled. It is likely that the estimated density of posterior probability distributions is low when concept drift occurs. Figure 1 shows the estimated density distributions before and during concept drift occurs. When concept drift occurs, the estimated densities concentrate around zero whereas distribution before concept drift occurs is widely spread.

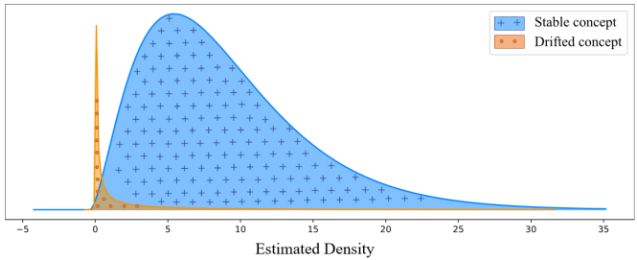

Figure 1: Example distributions of estimated densities for Hyper-Plane dataset (Bifet et al. 2010).

Primary contributions of this paper are as follow:

1) A semi-supervised framework is proposed that can learn and adapt well in a streaming environment with the presence of real concept drift where there is little to no data labels available.

2) The framework has the ability to detect the real concept drift occurring under such conditions.

3) A novel concept drift detection technique is developed to compare posterior probability distributions for partially labeled data streams based on density estimation. The intuition of developing this concept drift detection technique is that conventional methods for comparing distributions are based on the statistical information of the distributions. This is unreliable for distributions generated from partially labeled data.

## Related Work

Supervised approaches for concept drift detection often require full availability of labeled data to compute performance metrics such as accuracy, error rate, F-measure, precision and recall which are sequentially monitored as a signal for a change in concept (Hu, et al., 2019).

Supervised state of art concept drift detection methods such as Drift Detection Method (DDM) monitor the error probability, $p_i$ and standard deviation $s_i$ over time (Gama, et al., 2004). The condition $p_i + s_i > p_{min}+3s_{min}$ where $p_{min}$ and $s_{min}$ are the record minimum values signifies that the performance distribution of the current data deviates from historical record. Baena-García, et al. (2006) extends this idea into Early Drift Detection Method (EDDM) with the aim to detect gradual drift where DDM failed. EDDM monitors the distance between two subsequent errors as a drift detection metric. For a stable concept, it is assumed that the distance is large between the two errors. An ADaptive sliding WINdow (ADWIN) approach was developed by Bifet & Gavaldà (2009) that compares the mean performance metrics of recent data with the historical data. Window size of recent data shrinks when there is a significant difference between both means and vice versa. Page Hinckley Test (PH Test) monitors the difference between the cumulative means of the observed measurement and the minimum cumulated mean to see if it significantly deviates from zero (Mouss, et al., 2004).

Semi-supervised approaches learn and adapt to data streams with the presence of concept drift by exploiting the limited labeled portions of the data. A Semi-supervised Adaptive Novel class Detection (SAND) framework utilizes the confidence scores of a trained classifier on incoming unlabeled data as a drift detection measurement (Haque, et al., 2016). The use of confidence scores becomes unreliable if data attributes distribution remains while class distribution changes. Kmieciak & Stefanowski (2011) proposes an approach to handle sudden concept drift in data attributes distribution. Hosseini, et al. (2016) proposed Semi-supervised Pool and Accuracy based Stream Classification (SPASC) which claims to be able to deal with recurring drift situation. A new classifier is built when performance of all classifiers in the ensemble failed to achieve certain performance. Hence when a recurring concept drift occur, this method can adapt to it using the ensembles of classifiers. It is not suitable to handle a new and non-recurring drift. Qin & Wen (2018) proposes similar approach to SPASC by modifying the replacement method. These methods are different from the focus of our proposed framework which is to detect real drift in a streaming environment with limited access to labeled data.

Unsupervised approaches to detect concept drift focuses on virtual drift. dos Reis, et al. (2016) developed an incremental Kolmogorov-Smirnov Test (KS Test) to measure the dissimilarity of current and historical data attribute distribution in a data stream. de Mello, et al. (2019) uses Uniform Stability, a mathematical function to determine the stability of data attribute distribution. Another unsupervised method defined a new measurement called Unified Strangeness Measure (USM) which determines how much a data point is different from others (Mozafari, et al., 2011). This measure is then used to compute martingale value and compare it to a user defined threshold for drift detection.

## Problem Definition

Assume a data stream $D$ is given with a set of input variables $X \in R^d$, where $d$ is the number of features. The given data stream can be divided into $W$ windows of instances of size $n$. The target variable $y$ of the input variables $X$ within $W$ is either labeled $L$ or unlabeled $U$. The task is to detect the change in probability of $y$ given $X$ i.e., $p(y/X)$. Conventional methods to detect changes in $p(y/X)$ require all $y$ in $W$ to be fully labeled. These methods store a historical $p(y/X)_{old}$ as a reference to be compared with the current $p(y/X)$. In a streaming environment, $W$ comes in high velocity and volume. It is impractical to either assume that all $y$ in $W$ are $L$ or to store all $X$ for further processing due to memory limit. Hence, we need a method to detect the changes in $p(y/X)$ distribution with only limited $L$ available in $W$ without storing any historical $p(y/X)_{old}$ for reference.

The proposed framework is able to deal with the shortcomings of conventional methods as mentioned above. To learn from limited $L$ in $W$, we introduce a knowledge discovery component in the framework which incorporates different learning methods that discover reliable labeled data (called $RL$), from the unlabeled portion, $U$ of $W$.

**Definition 1:** *Reliable labeled (RL) data – The estimated data labels inferred (or extracted) for the unlabeled data with high confidence.*

The framework detects changes in $p(y/X)$ by monitoring the posterior distributions directly from $RL$. We developed a novel method which utilizes the densities of $p(y/X)$ distributions instead of its statistical information to detect real drift. To deal with the memory limitation, instead of storing the reference $p(y/X)_{old}$ for comparison, the proposed framework represents the current and previous concepts using two different estimators.

## Proposed Framework

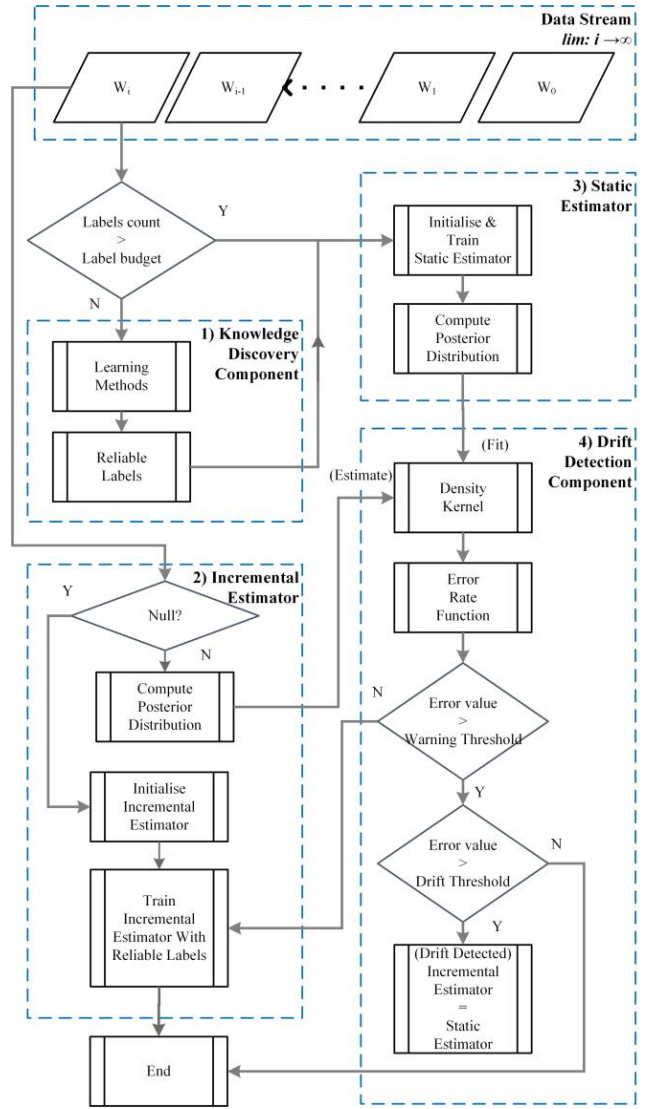

Figure 2: High level workflow of proposed framework for one data window.

The proposed framework can be separated into four components i.e., 1) knowledge discovery (KD) component, 2) incremental estimator, 3) static estimator, and 4) drift detection component. **Knowledge discovery (KD) component** aims to discover the data labels from unlabeled data stream using various learning methods to obtain portions of labels depending on the labeling budget allowed. **Incremental estimator** acts as a base learner in the framework to incrementally learn from the reliable labeled data from KD where the posterior probabilities from the incremental estimator represents the concepts of previous data window. **Static estimator** is initialized and trained directly from reliable labeled data for every data window. The posterior distribution computed from the static estimator represents the concepts of the incoming data window. The **drift detection component** employs our novel drift detection technique developed which utilizes the density of posterior distributions from both incremental and static estimators. The intuition of adopting posterior distributions densities is to detect real concept drift. An error rate function is used to determine the diffusion of both distributions. When the error rate value drops below a specified drift threshold, concept drift is detected. In order to detect gradual drift, a warning threshold is used to stop the base estimator from learning incrementally. This is to further verify if a concept drift is about to happen or it is just a weak estimate of the density. If a gradual drift is about to happen, the error rate value will continue to drop until the drift threshold is reached. This is shown clearly in figure 4 in the later section. Figure 2 illustrates the high-level workflow while algorithm 1 describes the process of the proposed framework.

### Knowledge Discovery Component

Labels are necessary to estimate the real concepts from data streams. Without labels, concepts are merely just a representation of the data attributes distribution which does not capture the relationship between the class and data attributes.

In this framework, reliable labeled data are extracted from unlabeled or partially labeled data stream. Various learning method can be used to obtain reliable labels from a data stream. Obtaining data labels are expensive. Depending on the labeling budget available, this framework only extracts the portion of labeled data allowed. In this paper, Active learning and Positive Unlabeled Learning (PU Learning) are explored as the learning methods to estimate labels in the knowledge discovery component.

#### A) Active Learning

Active learning method is a selective labeling technique which selectively requests for labels in a data stream. The percentage of the labels requested depends on the system's allowance for the percentage labeling budget.

In this paper, a random selective labeling technique is employed as Žliobaitė, et al. (2013) has shown the effectiveness of this technique. The random selective labeling technique randomly selects percentage of instances in the data window to be labeled. This technique is simple and effective as the selective labeled data are uniformly distributed in the data space hence avoiding bias sampling. In this paper, we assume that label budget is greater than 0.

#### B) Positive Unlabeled Learning

In certain cases, the data streams arrive with only partial positively labeled data. As positive labels are already available, PU learning method is used to extract the reliable negative data (Li, et al., 2009).

It is worth noting that the unlabeled portion of the data is a mix of negative and positive instances. Our proposed framework employs the biased learning technique (Bekker & Davis, 2018) to extract reliable negative data with random sampling technique similar to Active learning method. The biased learning method treats all unlabeled data as negative data and trains a classifier with the data. While only a certain percentage of the positive data are labeled, we randomly draw the same percentage of the positive data from the negative data sample to obtain uniformly distributed negative instances to avoid bias sampling.

**Algorithm 1.** Semi-supervised concept drift detection algorithm

1. Input: $n$ //window size, $B_L$ //label budget,
2.     $L$ //labeled instance, $U$ //unlabeled instance,
3.     $W \in \{(X_0, y_0 \in \{L, U\}) \ldots (X_n, y_n \in \{L, U\})\}$
4.     $i\_clf \leftarrow$ Null //incremental estimator,
5.     $\tau$ //drift threshold, $\varphi$ //warning threshold
6. **while** True **do**
7.    obtain $W$ from data stream, $D$
8.    $RL \leftarrow \{\}$ //Initialize empty set for reliable labels
9.    **if** $W[y == L]$.count()$/n < B_L$ **do**
10.      $RL \leftarrow KD(W)$ // Obtain reliable label
11.    $RL \leftarrow RL \cup W[y == L]$
12.    $s\_clf \leftarrow$ HoeffdingTree() //static estimator
13.    $s\_clf \leftarrow s\_clf$.train($RL$) //train static estimator
14.    $s\_p \sim N(0, 1) \leftarrow post\_proba(RL.X, RL.y)$
15.    $k\_e \leftarrow$ KernelDensity.fit($s\_p$) //fit density kernel
16.    **if** $i\_clf$ is Null **do**
17.      $i\_clf \leftarrow$ HoeffdingTree() //incremental estimator
18.      $i\_clf \leftarrow i\_clf$.train($RL$) //train incremental estimator
19.    **else do**
20.      $i\_y \leftarrow i\_clf$.predict($W$)
21.      $i\_p \sim N(0, 1) \leftarrow post\_proba(W, i\_y)$
22.      $\rho \leftarrow k\_e$.estimate($p2$) //estimate density
23.      $\rho \leftarrow$ scale[0, 50 * $e^{-4\rho} + \delta$] //sensitivity control
24.      $\varepsilon \leftarrow erf(\rho)$ //error rate function
25.      **if** $\varepsilon < \varphi$ **do** //below warning threshold
26.        **if** $\varepsilon < \tau$ **do** //below drift threshold
27.          $i\_clf \leftarrow s\_clf$ //replace incremental estimator
28.      **else do**
29.        $i\_clf \leftarrow i\_clf$.train($RL$)

### Incremental Estimator

The incremental estimator in the framework learns incrementally from the reliable labeled data extracted from the knowledge discovery component. Before this estimator learns incrementally from the current reliable labeled data, the posterior probabilities are first estimated by predicting the target variable $y$ for the incoming data window. Hence this estimated posterior probability distribution represents the concepts of previous data window.

### Static Estimator

Static estimator in the framework trains from the current reliable labeled data. The proposed framework represents the concepts of current data window using the posterior probability distribution computed from the current reliable labeled data. These estimators can be any learner that is able to learn well and incrementally with the type of data in the stream. As the static estimator acts as a replacement backup for the incremental estimator, both estimators chosen should have similar classification performance on a given type of data.

## Drift Detection Component

Unlike most semi-supervised state-of-the-art methods which focus on the change in data attribute distribution *p(X)*, our proposed framework utilizes the limited labeled data from knowledge discovery component to compute the posterior probabilities for drift detection.

### A) Posterior Probability Distribution

Posterior probability for a targeted variable *y* given an input variable *x* is:

$$p(y|x_i) = \frac{p(x_i|y) \times p(y)}{p(x_i)} \quad (1)$$

The *y* in the equation represents the target variable that is being monitored where $x_i$ is one of the data attributes from the input variable *X*. Approaches that detect virtual drift compute *p(X)* directly from the input variable *X*. These approaches ignore the changes in *p(y)*. Hence, detecting virtual drift which is also the change in *p(X)* does not necessarily capture the change in *p(y)*. The change in *p(y)* is known as the change in prior belief which is important as there are many other hidden contexts which are usually uncaptured in the data attributes affecting the learning tasks. Our framework monitors the change in the posterior distribution *p(y/X)* to capture the overall change in the real concept.

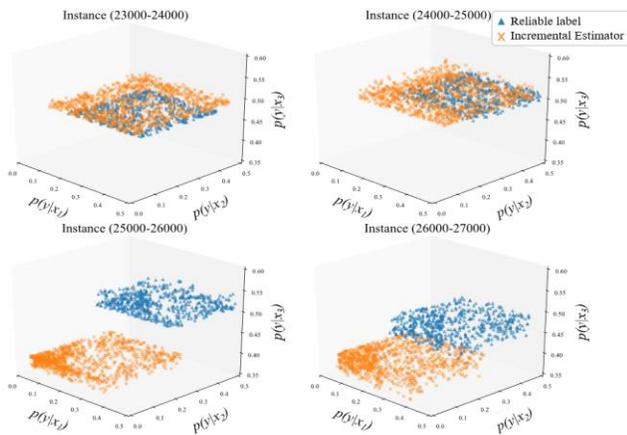

Figure 3: Example of sequential posterior probabilities shift of SEA dataset with 60% labels. Blue – Distribution of reliable labels; Orange – Distribution of incremental estimator.

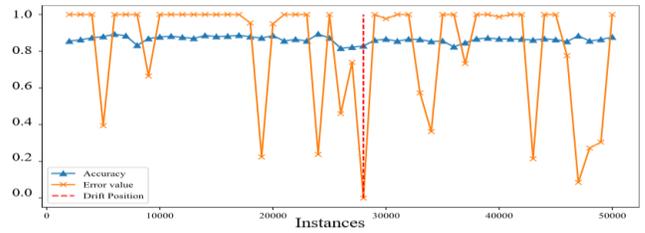

Figure 4: Accuracy of incremental estimator and error rate value from framework with drift threshold of 0.05 of SEA dataset (Bifet et al. 2010) with 60% labels.

In the proposed framework, for each estimator, the posterior probabilities of the target class of each data instance in the data window with respect to each attribute are computed. For example, $p(y|x_1)$, $p(y|x_2)$ and $p(y|x_3)$ are computed for a data stream with three features which are then used for density estimation in drift detection component. An example of posterior distribution shifting sequentially is depicted in figure 3. When concept drift occurs at the instance index of 25000, there is a significant difference between both distribution of reliable labels and incremental estimator's which result in a low estimated density. This can also be seen from figure 4 which shows the graph of the error rate value, accuracy of incremental estimator, and the position where drift is detected. There are some uncertainties on the error rate as this example only uses 60% of the labels in the data.

### B) Density Estimation

As the two distributions are computed from different estimators with varying sample size, statistical comparison such as Kolmogorov-Smirnov Test and T-test of the two distributions are often too sensitive and unstable. Another reason that statistical test deemed to be unstable in our case is because only labels in the data window are used. Hence, many uncertainty areas are presented in the data space.

To overcome this challenge, the proposed framework employs kernel density estimation to estimate the density of the posterior probabilities of classes to each data attributes. The kernel is fitted with the reliable labeled data's posterior probabilities to predict the overall densities of the incremental estimator's posterior probabilities.

Number of false alarms are inversely correlated to the label percentage. False alarm is defined as the drift detected at the incorrect instance location. Hence, a scaling factor as shown in equation 2 is used to control the sensitivity. This equation describes the relationship between the number of false alarms and the label percentage.

$$\gamma = 50 \times e^{-4\alpha} + \delta \quad (2)$$

γ in equation 2 is the scaling factor and α represents the label percentage available in the dataset. δ is the parameter that controls the overall sensitivity of the framework. As illustrated in figure 5, fewer false alarm were detected after applying the scaling factor to the estimated density distribution.

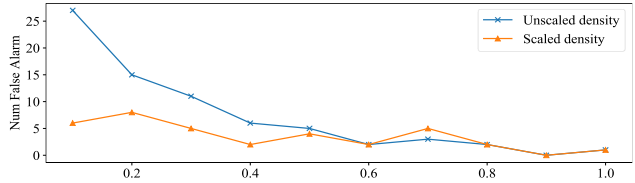

Figure 5: Relation of false alarm and label percentage.

**C) Error rate function**

The proposed framework detects concept drift by measuring the diffusion of the estimated density. The diffusion measurement can be described by equation 3 which is also function of error rate with an output value of [0, 1] where 0 indicates concept drift and 1 indicates stable concept.

$$erf(x) = \frac{1}{\sqrt{\pi}} \int_0^x e^{-t^2} dt \quad (3)$$

This error rate function is a sigmoid function for probability and statistic distribution to describe diffusion of the data (Chevillard, 2012). The x in equation 3 used in the proposed framework is the average density.

# Evaluation

## Experimental Setup

The aim of this experiment is to show that in circumstances with only small percentage of labels are available, the framework is able to detect concept drift and achieve comparable classification performance to state-of-the-art methods. We compare the proposed framework with several supervised state-of-the-art drift detection techniques i.e. DDM (Gama, et al., 2004), EDDM (Baena-García, et al., 2006), ADWIN (Bifet & Gavaldà, 2009), and PH Test (Mouss, et al., 2004). These techniques were selected as the type of drift to be detected are the same (real drift).

In our experiments, data window is set consistently as 1000 instances and the label percentage varies from 20% to 100%, Active learning is employed as the knowledge discovery component, Hoeffing Tree (Domingos & Geoff, 2000) as the incremental and static estimator, and posterior probabilities density estimation as drift detection component with 0.05 as drift threshold which signifies that we are 95% confident if a drift is detected.

For the state-of-the-art methods, Hoeffding Tree was also used as the base estimator. Similar adaptation strategy is applied across the experiments where a new estimator is built by training from the recent incoming data when a warning is signaled. When concept drift is signaled, the base estimator is replaced with the newly trained estimator to quickly adapt to the drift and to minimize classification performance loss.

[1] https://moa.cms.waikato.ac.nz
[2] https://archive.ics.uci.edu/ml/index.php

| Dataset Name | Num of Instances | Num of Classes | Num of Attributes |
|---|---|---|---|
| HyperPlane | 100,000 | 2 | 10 |
| SEA | 100,000 | 2 | 3 |
| ForestCover | 100,000 | 7 | 10 |
| Elec | 45,312 | 2 | 5 |

Table 1: Summary of datasets characteristics.

## Datasets

We have used four datasets in our experiments: two synthetic and two real datasets as depicted in Table 1. HyperPlane and SEA are synthetic datasets generated from MOA[1] framework. ForestCover is a real-world dataset that describes the forest cover type obtained from UCI repository[2]. The qualitative data attributes were removed to simplify the classification process. Elec[3] is a real-world dataset describes the electricity demand in New South Wales, Australia. The data attributes with missing data were removed from the dataset as our framework does not deal with missing data. Concept drifts were simulated in synthetic datasets to investigate how well the proposed framework detects drift compared to state-of-the-art methods. The real-world datasets were used to test how well the proposed framework perform and adapts to changing situations compared to state-of-the-art methods.

## Experimental Results

Table 2 is a summary of the experimental results. In this experiment, we compared the overall classification performance of the proposed framework along with the state-of-the-art methods. We named our framework as "(*% label*) DensityEst" in this section where different percentages of labels were experimented. Average accuracy throughout the experiment is compared to show that the proposed framework's performance is comparable to state-of-the-art methods. However, it is important to note that average accuracy does not determine the ability to detect concept drift. Bifet (2017) has shown that the accuracy of an incremental classifier which does not detect drifts achieves higher performance in terms of accuracy compared to classifiers equipped with drift detection methods.

We are also interested in the ability of the framework to detect concept drift compared to state-of-the-art methods. For the Hyperplane data set, concept drift is simulated at the instance index 75000. Three concept drifts are simulated in the SEA dataset which are at the instance indices of 25000, 50000 and 75000.

[3] http://www.inescporto.pt/~jgama/ales/ales_5.html

| Datasets → | HyperPlane | | SEA | | ForestCover | | Elec | |
|---|---|---|---|---|---|---|---|---|
| (% label) Drift Detection Method | Avg Accuracy | Num of Drift | Avg Accuracy | Num of Drift | Avg Accuracy | Num of Drift | Avg Accuracy | Num of Drift |
| PH | <u>0.829</u> | 13 | <u>0.874</u> | 4 | <u>0.805</u> | 24 | 0.743 | 14 |
| ADW | **0.833** | 18 | 0.869 | 8 | 0.791 | 44 | <u>0.756</u> | 28 |
| EDDM | 0.777 | 8 | 0.871 | 8 | 0.794 | 121 | 0.752 | 170 |
| DDM | 0.754 | 5 | 0.869 | 2 | 0.783 | 162 | 0.718 | 99 |
| (1.0) DensityEst | 0.826 | 8 | **0.876** | 4 | **0.808** | 9 | 0.755 | 10 |
| (0.9) DensityEst | 0.828 | 6 | <u>0.874</u> | 4 | 0.795 | 25 | 0.745 | 12 |
| (0.8) DensityEst | <u>0.829</u> | 5 | 0.871 | 7 | 0.794 | 15 | 0.738 | 13 |
| (0.7) DensityEst | 0.822 | 8 | 0.865 | 6 | 0.786 | 18 | 0.756 | 13 |
| (0.6) DensityEst | 0.824 | 10 | 0.867 | 4 | 0.788 | 29 | 0.739 | 15 |
| (0.5) DensityEst | 0.825 | 11 | 0.867 | 4 | 0.790 | 27 | **0.767** | 9 |
| (0.4) DensityEst | 0.822 | 7 | 0.868 | 5 | 0.767 | 34 | 0.732 | 16 |
| (0.3) DensityEst | 0.828 | 9 | 0.861 | 7 | 0.778 | 21 | 0.742 | 13 |
| (0.2) DensityEst | 0.824 | 12 | 0.861 | 5 | 0.759 | 29 | 0.745 | 12 |

Table 2: Average classification accuracy & number of drifts detected. Best performance is highlighted; second best is underlined.

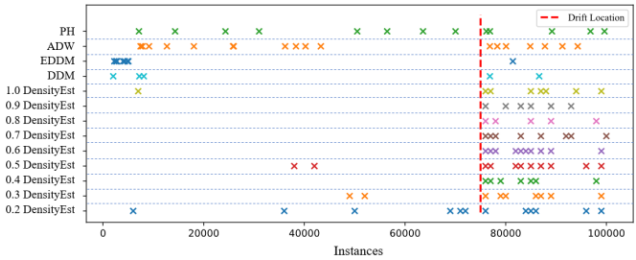

Figure 6: Drifts detected on HyperPlane Dataset.

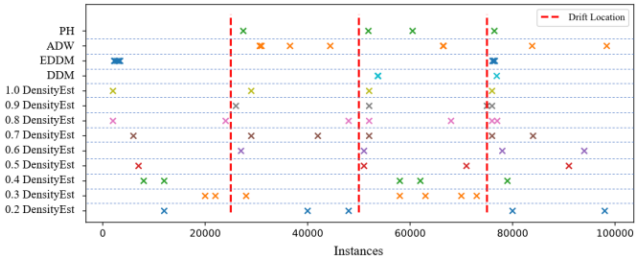

Figure 7: Drifts detected on SEA Dataset.

Figures 6 and 7 depict the comparison of detected drift positions of different methods with the drift location indicated with the red dotted line on Hyperplane and SEA datasets respectively. Our proposed framework is able to detect concept drift in both synthetic datasets with a few false alarms at lower percentages of label availability. EDDM and some low percentages labels in DensityEst did not perform as well on SEA dataset while other methods did detect the drift with a few false alarm or delayed detection. Most methods are able to detect the drift in HyperPlane dataset but our proposed framework results in fewer false alarms at the area before drift occurs. This shows that even with partially labeled data, it is possible to detect real concept drift while achieving comparable classification performance to state-of-the-art methods.

## Conclusion

In this paper, we made the following contributions to address this major problem:
1) A semi-supervised framework that learns and adapts well to data stream with low availability of labelled data and the presence of real concept drift.
2) Our framework detects real concept drift under low availability of labelled data by directly monitoring the change in posterior probability distribution over time.
3) A novel real drift detection mechanism was developed which monitors the overlapping density areas of the posterior probabilities from two estimators.

We showed that even with partially labeled data, it is possible to detect real concept drift while achieving comparable classification performance to state-of-the-art supervised methods. The purpose of the proposed framework is to deal with challenges of limited availability to label information in data streams and the need to address the real concept drift under such circumstances rather than to improve the state-of-the-art methods' classification performance.

Future research can make use of the posterior probability distribution of each class with respect to each data attribute to develop an inference-based system that provides information regarding which data attributes is causing the concept drift. Ensembles of classifiers can be employed in future research to investigate how well the framework is able to handle different concept drift types (e.g. recurring, gradual and abrupt drift). Similar techniques to ADWIN can be done to adapt the window size along with performance where Hoeffding bound (Phillips, 2012) is used.